\documentclass[sigconf]{acmart}
\renewcommand\footnotetextcopyrightpermission[1]{} 
\setcopyright{none}

\AtBeginDocument{%
  \providecommand\BibTeX{{%
    \normalfont B\kern-0.5em{\scshape i\kern-0.25em b}\kern-0.8em\TeX}}}

\setcopyright{acmcopyright}
\copyrightyear{2024}
\acmYear{2024}
\acmDOI{XXXXXXX.XXXXXXX}
\acmConference[CIKM'24]{Conference on Information and Knowledge Management}{October 21 – 25, 2024}{Boise, Idaho, USA}
\acmISBN{978-1-4503-XXXX-X/18/06}

\usepackage{colortbl}
\usepackage{multirow}
\usepackage{makecell}
\usepackage{graphicx}
\usepackage{subcaption}
\usepackage{booktabs}
\usepackage{caption}
\usepackage{hhline}
\usepackage{tabularx}
\newcommand{\modelName}{LVLM4EV}
\newcommand{\FNmodelName}{LVLM4FV}
\newcolumntype{Y}{>{\centering\arraybackslash}X}

\begin{document}

\title{Multimodal Misinformation Detection using Large Vision-Language Models}

\author{Sahar Tahmasebi} \email{sahar.tahmasebi@tib.eu} \orcid{0000-0003-4784-7391} \affiliation{ \institution{TIB – Leibniz Information Centre for Science and Technology} \city{Hannover} \country{Germany}}

\author{Eric Müller-Budack}
\email{eric.mueller@tib.eu}
\orcid{0000-0002-6802-1241}
\affiliation{ \institution{TIB -- Leibniz Information Centre for Science and Technology; \\ L3S Research Center, Leibniz University Hannover}
\city{Hannover} 
\country{Germany}}

\author{Ralph Ewerth}
\email{ralph.ewerth@tib.eu}
\orcid{0000-0003-0918-6297}
\affiliation{ \institution{ TIB -- Leibniz Information Centre for Science and Technology; \\  L3S Research Center, Leibniz University Hannover}
\city{Hannover}
\country{Germany}}

\renewcommand{\shortauthors}{Tahmasebi et al.}

\begin{abstract}
The increasing proliferation of misinformation and its alarming impact have motivated both industry and academia to develop approaches for misinformation detection and fact checking. Recent advances on large language models~(LLMs) have shown remarkable performance in various tasks, but whether and how LLMs could help with misinformation detection remains relatively underexplored. 
Most of existing state-of-the-art approaches either do not consider evidence and solely focus on claim related features or assume the evidence to be provided. Few approaches consider evidence retrieval as part of the misinformation detection but rely on fine-tuning 
models. In this paper, we investigate the potential of LLMs for misinformation detection in a zero-shot setting. 
We incorporate an evidence retrieval component into the process as it is crucial to gather pertinent information from various sources to detect the veracity of claims. 
To this end, we propose a novel re-ranking approach for 
multimodal evidence retrieval using both LLMs and large vision-language models (LVLM). 
The retrieved evidence samples~(images and texts) serve as the input for an LVLM-based approach for multimodal fact verification~(\FNmodelName). 
To enable a fair evaluation, we address the issue of incomplete ground truth for evidence samples 
in an existing evidence retrieval dataset by annotating a more complete set of evidence samples for both image and text retrieval. Our experimental results on two
datasets demonstrate the superiority of the proposed approach in
both evidence retrieval and fact verification tasks and also better
generalization capability across dataset compared to the supervised
baseline.
\end{abstract}



\maketitle


\section{Introduction}
\begin{figure}[t]
  \centering
  \includegraphics[width=\linewidth]{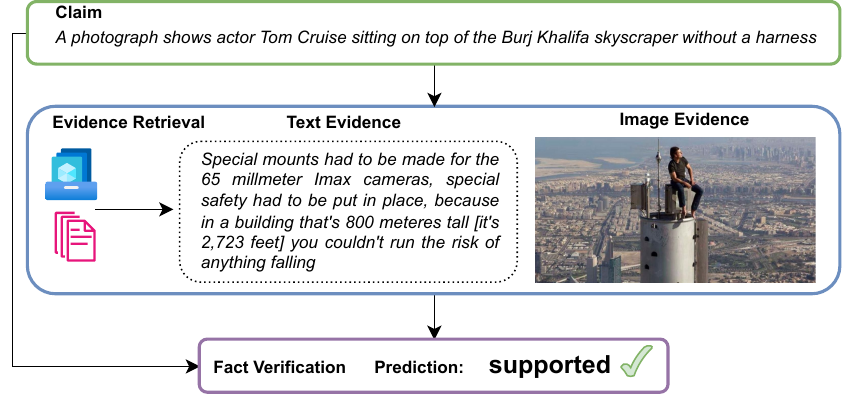}
  \caption{Example of multimodal misinformation detection.}
  \label{teaser_fig}
\end{figure}

Misinformation and fake news contain false information to deliberately deceive readers~\cite{allcott2017social} and have become a pressing challenge since they can cause severe consequences for society.
Typically, misinformation is conveyed in different modalities such as images, text, and videos to provide a stronger story line and attract attention from readers.
The situation has become even more complicated with the emergence of large language models~(LLMs) like Generative Pre-trained Transformer~(GPT)~\cite{DBLP:GPT4}, 
since they can be intentionally misused to generate~\cite{DBLP:chatgpt-missuse} or spread misinformation due to the hallucination issue~\cite{DBLP:chatGPT-diagnostic}. 
Thus, automated solutions for fact checking and multimodal misinformation detection are in great need.
The multimodal misinformation detection is generally structured into distinct stages including claim detection and extraction, check-worthiness prediction, verdict prediction and explanation generation. 
This work focuses on the critical stage of verdict prediction which comprises an evidence retrieval component followed by the prediction of the targeted claim regarding its truthfulness~(see Figure~\ref{teaser_fig}).

Recently, researchers have started to investigate automatic misinformation detection by developing various benchmark datasets~\cite{nakamura-etal-2020-fakeddit, DBLP:fever, DBLP:mocheg}. State-of-the-art approaches~\cite{DBLP:Tahmasebi,DBLP:mocheg,DBLP:EANN} have mainly used deep learning techniques to extract features from text and image. 
However, detecting misinformation on (multimodal)~social media is a challenging research problem since it is difficult to maintain applicability and good performance for unforeseen events. 
We found the following limitations with the current studies. (1)~Most of them only consider text~\cite{shu2017fake, DBLP:Liar, DBLP:conf/icwsm/BozarthB20} while ignoring the multimodal nature~(e.g., combination of images and texts) of online articles, which are useful to predict the truthfulness of claims. 
(2)~While evidence retrieval is a core sub-task in detecting misinformation, most multimodal-based approaches either do not use evidences and rely only on the claim-related features from the associated news~\cite{DBLP:BDANN, DBLP:EANN,DBLP:Tahmasebi}, or assume that evidences are already provided~\cite{DBLP:Liar, DBLP:logically} based on which the models can directly predict the truthfulness of the target claim. 
However, this is not realistic in practice as the claim does not typically come with evidence. Instead evidences should be retrieved from a knowledge base or the Internet.
(3)~Only few approaches~(e.g., \cite{DBLP:mocheg}) consider evidence retrieval as part of multimodal misinformation detection. However, these approaches have not leveraged powerful generative LLMs~(e.g.,~\cite{DBLP:LLAMA, DBLP:Mistral}) and LVLMs~(e.g.,~\cite{DBLP:Instruct-blip,DBLP:llava} for retrieval and the classification of claims regarding misinformation. 
%
%
%
These LLMs and LVLMs 
have demonstrated superior comprehension capabilities in tasks like image captioning and visual question answering. Their strong zero-shot and few-shot generalization opens doors to myriad of applications across many domains.

In this paper, we introduce a pipeline for multimodal misinformation detection including a novel re-ranking method for evidence retrieval using LLMs and LVLMs followed by a fact verification step. 
Overall, we make the following contributions. 
%
(1)~In contrast to related work that either does not consider evidence~\cite{DBLP:BDANN,DBLP:Tahmasebi,DBLP:EANN} or assume it to be provided~\cite{DBLP:Liar, DBLP:conf/emnlp/ZlatkovaNK19,DBLP:logically}, 
we incorporate an evidence retrieval component into our misinformation detection approach, to make an informed decision regarding the veracity of claims.
(2)~Unlike other approaches that consider evidence retrieval as part of multimodal misinformation detection using supervised  models, we propose a novel unsupervised re-ranking approach 
leveraging LLMs and LVLMs for evidence retrieval called \modelName. 
For this purpose, we propose effective prompting strategies and extract ranking scores directly from both LLMs and LVLMs for text and image retrieval. 
The retrieved image and text evidences serve as input for fact verification using L(V)LMs as a classifier through a prompting strategy 
that classifies misinformation using majority voting on the answers. An example is shown in Figure~\ref{teaser_fig}.
(3)~
Existing datasets~\cite{DBLP:mocheg,DBLP:factify} create ground-truth evidences for a claim by retrieving (single) expert-verified evidences from sources such as \textit{PolitiFact}\footnote{https://www.politifact.com} or \textit{Snopes}\footnote{https://www.snopes.com}. 
However, in a large database, this might overlook other pertinent evidence and thus some relevant evidences for a claim remain unlabeled. 
We identify this 
issue of \textit{incompelete ground-truth labels} in the \textit{MOCHEG} dataset for the evidence retrieval task 
and resolve it by annotating a 
more complete evaluation subset, consisting of top-10 relevant candidate evidences for both image and text retrieval to ensure fair and accurate assessment of system performance.
(4)~
We perform experiments on two datasets to evaluate the effectiveness of the proposed pipeline and also explore generalization capability across datasets, which is critical for real-world applications. The findings underscore the effectiveness of our unsupervised, multimodal approach for misinformation detection and an improved generalization capability compared to supervised baselines.

The remainder of this paper is structured as follows. Section~\ref{sec:related_work} reviews related work on misinformation detection and large generative AI models. Our pipeline for multimodal misinformation detection, including a novel re-ranking approach based on LLM and LVLM for evidence retrieval, as well as an approach for fact verification is presented in Section~\ref{sec:methodology}. The experimental setup and results including details of annotation for a clean evaluation set to address incomplete ground truth issue are presented in Section~\ref{sec:experiments}.
Section~\ref{sec:conclusion} concludes this paper and provides future work directions.
\begin{figure*}[t]
  \centering
  \includegraphics[width=0.93\linewidth]{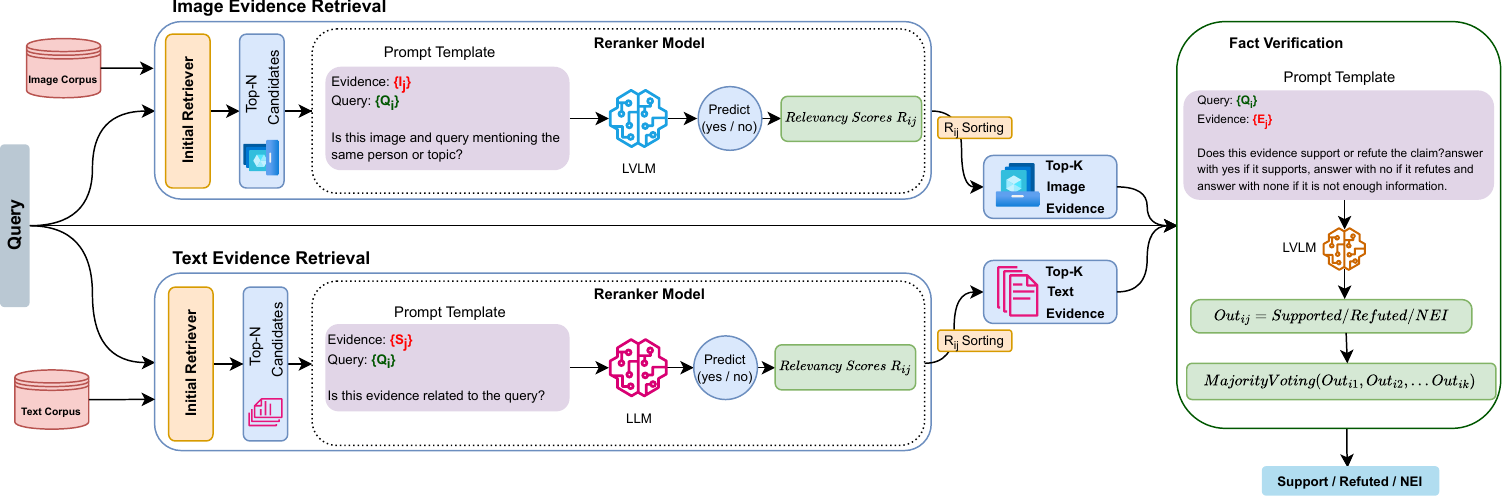}
  \caption{Overview of our misinformation detection approach. Blue border~(\modelName): Based on a textual input claim, evidence texts and images are initially retrieved from a corpus using a state-of-the-art approach~(e.g.,~\textit{MOCHEG}~\cite{DBLP:mocheg}). 
  Generative LLMs~(e.g.,  Mistral-7B~\cite{DBLP:Mistral}) and LVLMs~(e.g., InstructBLIP~\cite{DBLP:Instruct-blip}) are used to re-rank the top-N text and image evidences.
  Green border~(\FNmodelName): Based on the re-ranked evidences, we finally employ a LVLM~(e.g.,~LLaVA~\cite{llava-mistral}) for misinformation detection.}
  \label{fig:pipeline}
\end{figure*}
\section{Related Work}
\label{sec:related_work}
In this section, we review the related work on misinformation detection approaches, including pattern-based and evidence-based methods, as well as large generative AI models.
\subsection{Misinformation Detection}
\label{sec:rw_misinformation}
Misinformation detection methodologies can be divided into two main categories: pattern-based and evidence-based~\cite{DBLP:EANN}.

\textbf{Pattern-based methods} treat the misinformation detection as a feature classification task, where language/vision models are employed to determine the integrity of news content. This evaluation encompasses various aspects such as writing style, sentiment analysis, user credibility and other relevant features.
Majority of previous research in this category heavily depends on features from text and user metadata~\cite{DBLP:conf/acl/SteinPKBR18, DBLP:conf/semweb/PanPLLLL18}. Recent research has begun incorporating images~\cite{DBLP:BDANN,DBLP:MediaEval2015,DBLP:Tahmasebi, DBLP:InfoSurgeon} and videos~\cite{DBLP:conf/icwsm/MicallefSCAKM22,DBLP:journals/oir/PapadopoulouZPK19} into the detection of misinformation due to the inherent multimodality of information. Most of these approaches rely on verifying cross-modal consistency~\cite{DBLP:journals/ipm/SongNZW21,DBLP:conf/emnlp/TanPS20} or generating a combined representation of textual and visual features for classification~\cite{DBLP:Tahmasebi, DBLP:BDANN,DBLP:journals/ieeemm/KambojHARA21}. For example, Wang et al.~\cite{DBLP:EANN} introduced an \textit{Event Adversarial Neural Network~(EANN)}, which extracts event-invariant features to detect fake news on unforeseen events. From the same perspective, \textit{Spotfake}~\cite{DBLP:spotfake} combined features from both modalities and showed the effectiveness of pre-trained language models like \textit{BERT}~\cite{DBLP:BERT} and computer vision approaches like Visual Geometry Group model~(\textit{VGG-19)}~\cite{DBLP:VGG19}. 

\textbf{Evidence-based approaches} are designed to meticulously verify claims by integrating a wealth of external information. Early attempts on fact verification were based on textual information~\cite{DBLP:fever,DBLP:Liar,DBLP:defacto}. For example, The \textit{Liar} dataset and framework~\cite{DBLP:Liar} collected relevance evidences from \textit{PolitiFact} website and explored automatic fake news detection based on surface-level linguistic patterns. They proposed a hybrid convolutional neural network to integrate evidences with text and showed the improvement compared to evidence free deep learning model. Textual datasets and methods are no longer enough in the social media age as the claim and evidence could be in any modalities like image~\cite{chartcheck,DBLP:factify,DBLP:mocheg} or videos~\cite{DBLP:videoFC,DBLP:frenchvideoFC}. Gao et al.~\cite{DBLP:logically} framed the task as multimodal entailment task and proposed two baseline approaches including an ensemble model which combines two unimodal models and a multimodal attention network that models the interaction between image and text pair from claim and evidence document. Yao et al.~\cite{DBLP:mocheg} proposed an end-to-end Multimodal fact-CHecking and Explanation Generation~\textit{(MOCHEG)} benchmark dataset. To set the baseline for this benchmark, they have fine-tuned SBERT~(Sentence Bidirectional Encoder Representations from Transformers,~\cite{DBLP:SBERT}) for text retrieval and CLIP~(Contrastive Language–Image Pre-Training~\cite{DBLP:CLIP}), for image
retrieval with a contrastive loss. For the verification task, they used CLIP for claim and evidence embedding following an attention layer to compute distribution between them and a classification layer to get final labels.

While pattern-based approaches~(e.g.,~\cite{DBLP:BDANN,DBLP:Tahmasebi,DBLP:EANN}) do not consider retrieving evidences, most evidence-based approaches~(e.g.~\cite{DBLP:Liar, DBLP:conf/emnlp/ZlatkovaNK19,DBLP:logically}) operate under the assumption that relevant evidence is already provided alongside with the content under scrutiny. 
However, this assumption diverges from real-world scenarios since evidence is often required to verify claims but rarely provided.
Only few approaches~(e.g.,~\cite{DBLP:mocheg}) consider
evidence retrieval as part of multimodal misinformation detection and generally fine-tune pre-trained Small Language Models~(SLMs) like BERT~\cite{DBLP:BERT} to understand news content and provide fundamental representation. 
SLMs do bring improvements, but their knowledge and capability limitations also compromise further enhancement of misinformation detectors. For example, BERT was pre-trained on text corpus like Wikipedia and thus struggled to handle news items that require knowledge not included~\cite{DBLP:conf/cikm/ShengZCZ21}.
More powerful LLMs and LVLMs that have achieved impressive performance for many tasks, still remain to be explored  for the task of evidence-based misinformation detection.

\subsection{Generative AI Models}
\label{sec:rw_generative_ai}
%
%
Large Language Models (LLMs) such as GPT~\cite{chatgpt}, LLaMA~(Large Language Model Meta AI,~\cite{DBLP:LLAMA} and Mistral-7B~\cite{DBLP:Mistral} 
which are usually trained on the larger-scale corpus and aligned with human preferences, have shown impressive emergent abilities on various tasks~\cite{DBLP:journals/tmlr/WeiTBRZBYBZMCHVLDF22} and are considered promising as general task solvers. For example, in the similar task of fact checking, Hoes et al.~\cite{hoes2023leveraging} find that ChatGPT accurately classifies 69\% of text-only statements in a dataset built from the factchecking platform PolitiFact. \textit{FactLLaMA}~\cite{factllama} integrates external evidence into the instruct-tuning process to enhance the model’s ability to leverage evidence for predictions.
Hu et al.~\cite{DBLP:bad_actor} also studied how to utilize LLMs help in improving the performance of text-based fake news detection and showed that LLMs like \textit{GPT-3.5} underperforms the task-specific SLM but could provide informative rationales and complement SLMs in news understanding.

LLMs are expanding beyond text-only functions to include multiple modalities like image and videos. With the advancement of LVLMs like Bootstrapping Language-Image Pre-training~(BLIP-2)~\cite{DBLP:BLIP2}, \textit{InstructBLIP}~\cite{DBLP:Instruct-blip} and Large Language-and-Vision Assistant~(\textit{LLaVA})~\cite{DBLP:llava} the systematic evaluation of their capabilities has become increasingly critical. A broad array of benchmarks~\cite{DBLP:corr/abs-2306-09265, chartcheck,DBLP:corr/abs-2306-13549} has been established to estimate the LVLMs performance across a variety of tasks, such as image captioning, optical character recognition and chart fact checking. However, there is an absence of research on evaluating the ability of LVLMs to conduct multimodal misinformation detection.

%

\section{Multimodal Misinformation Detection}
\label{sec:methodology}

In this section, we propose a 
pipeline that exploits LLMs and LVLMs for multimodal misinformation detection. 
The task is to predict whether a post is \textit{refuted}~($y=0$), \textit{supported}~($y=1$) or the evidence provides not enough information \textit{NEI}~($y=2$). 
%
%
As illustrated in Figure~\ref{fig:pipeline}, it consists of two main components: (1)~An evidence retrieval component which automatically retrieves evidences from a large collection of web sources including articles and images for a given claim. (2)~A multimodal misinformation detection approach that predicts the veracity of the claim given the retrieved evidences. 

\subsection{Problem Definition}
We define the task of multimodal 
misinformation detection as follows. 
Given a textual claim~(or query)~$Q$ and a corpus with $m$evidences $\mathbb{C} = \{E_1, E_2, \dots,\ E_m\}$ as input, we aim to predict whether the claim is refuted ($y=0$), supported ($y=1$) or evidences do not provide enough information~(NEI, $y=2$). Each evidence~$E = (\mathbb{I},\mathbb{S})$ is defined as a set of $v$~images $\mathbb{I} = \{I_1, I_2, \dots,\ I_v\}$ and $t$~sentences $\mathbb{S} = \{S_1, S_2, \dots,\ S_t\}$. 


\subsection{Evidence Retrieval}
Most existing methods for misinformation detection prioritize claim verification and directly compute a fused representation of multimodal (textual and visual) information for final classification. 
However, retrieving high-quality evidences is the foundation of misinformation detection especially in terms of new events.

Given a textual claim $Q$ and a corpus of evidences $\mathbb{C}$ as input, the goal of initial text and image retrieval is to find a sorted subset of the $N$ most related sentences $\mathbb{S}' = \{S_1, S_2, \dots,\ S_N\} \subset \mathbb{C}$ and images $\mathbb{I}' = \{I_1, I_2, \dots,\ I_N\} \subset \mathbb{C}$. 
For both subtasks, we initially retrieve text and image evidences using an appropriate state-of-the-art model. 
Since existing supervised ranking methods can suffer from weak generalizability to new domains, and restricted commercial use~\cite{DBLP:journals/corr/abs-2112-09118}, we aim to exploit the excellent generalization capabilities of LLMs~\cite{beeching2023open} and LVLMs~\cite{DBLP:seed-bench2} in zero-shot settings to improve evidence retrieval.
To this end, we propose to use LLMs and LVLMs to re-rank text and images evidences, respectively.  In the following, we provide details for our proposed re-ranking approach, which we denote as \modelName~(Large Vision-Language Models for Evidence Retrieval).
%

\subsubsection{Initial Retriever}
\label{sec:initial retriever}
In information retrieval tasks, the utilization of an initial retriever is imperative due to the computational complexity of directly applying a computational-heavy model (especially LLMs or LVLMs) to the entire corpus~\cite{DBLP:Is_chatgpt}. 
The initial retriever efficiently narrows down the search space by quickly identifying a subset~$\mathbb{C}' \subset \mathbb{C}$ of potentially relevant evidences, ensuring that computational resources are utilized effectively while maintaining the performance.
As shown in Figure~\ref{fig:pipeline}, we employ an initial retrieval approach to retrieve text and image evidences from a large corpus. 
For thus purpose, we apply MOCHEG~\cite{DBLP:mocheg}, which is a state-of-the-art evidence retrieval approach, to get top-$N$ most related evidences. 

\paragraph{Text Retrieval}
For text evidence retrieval, documents are segmented into sentences, and the fine-tuned \textit{SBERT} model~\cite{DBLP:SBERT} from MOCHEG~\cite{DBLP:mocheg} is applied to generate contextual representations for input claims and sentences from the evidences. We compute the ranking score~$r_j = \mathtt{cossim}(\mathtt{SBERT}(Q), \mathtt{SBERT}(S_j))$ using the cosine similarity 
between the claim~${Q}$ and each sentences~$S_{j} \in \mathbb{C}$ from all evidences in the corpus to create an initial ranking and filter the top-$N$ evidences to output a sorted subset~$\mathbb{S}' = \{S_1, S_2, \dots,\ S_N\}$ used for re-ranking~(Section~\ref{ranking}).

\paragraph{Image Retrieval}
Similarly, for image evidence retrieval, we apply the text~$\mathtt{CLIP}^T$ and image encoder~$\mathtt{CLIP}^I$ from the CLIP model~\cite{DBLP:CLIP} fine-tuned from MOCHEG~\cite{DBLP:mocheg} to encode both textual input claim and images. We use the cosine similarity as ranking score $r_j = \mathtt{cossim}(\mathtt{CLIP}^T(Q), \mathtt{CLIP}^I(I_j))$ to rank all images~$I_{j} \in \mathbb{C}$ from each evidences in the corpus. 
As a result, a sorted subset of $N$~images $\mathbb{I}' = \{I_1, I_2, \dots,\ I_N\}$ is created for re-ranking~(Section~\ref{ranking}).
\subsubsection{Prompting}
\label{sec:prompting}
We use 
prompting to re-rank the results for both image and text retrieval. For this purpose, we use a prompt template to input the query claim~$Q$ and a initially retrieved sentence~$S \in \mathbb{S}'$ or image evidence~$I \in \mathbb{I}'$ into a generative AI model for relevance assessment. The models are instructed to output “Yes” if a candidate evidence is relevant to the claim and “No” otherwise. 
%
Given that prompt design significantly influences the response of an LLM or LVLM, different prompts have been used. Examples are shown in Figure~\ref{fig:pipeline} and are evaluated in the experiments.

\subsubsection{Re-ranking using Generative AI Models}
\label{ranking}
%
%


%
We intent 
to improve the original retrieved evidences by a re-ranking method~$f(\mathbb{C})' \mapsto \mathbb{C}''$ based on generative AI models, e.g., \textit{Mistral}~\cite{DBLP:Mistral} for text and \textit{InstructBLIP}~\cite{DBLP:Instruct-blip} for image evidences. 
To this end, we first apply the prompt templates mentioned in Section~\ref{sec:prompting} to the Top-$N$ text~$\mathbb{S}'$ and image evidences~$\mathbb{I}'$ from the initial retriever and the input claim~$Q$ to get an answer which is “Yes” or “No”. 
These answers then allow us to perform a re-ranking of the top-$N$ evidences~$\mathbb{C}'$. For this purpose, we consider the following strategies. 

\paragraph{Initial Ranking Scores~(\texttt{IRS})} In this approach, we re-rank evidences solely based on the answer of the LLM or LVLM. 
For this purpose, we rank all evidences that are considered relevant~(answer: “Yes”) by the LLM or LVLM higher than the irrelevant ones~(answer: “No”).
The remaining ranking, i.e., the ranking within relevant and irrelevant evidences, is created based on the ranking score~$r_j$ computed during the initial retrieval step~(Section~\ref{sec:initial retriever}) using the cosine similarity.  
Finally, we select the top-K re-ranked evidences from~$\mathbb{C}''$ as final evidence for fact verification~(Section~\ref{sec:fact_verification}).

\paragraph{Generative AI Scores~(\texttt{GAIS})} 
Each generative AI model uses a tokenizer for text generation with a specific dictionary of possible tokens. We use the log probability of the generative AI model to extract the scores for all pairs in order to re-rank them. 
For this purpose, we re-rank the evidences using the probability of model generating the token ‘Yes’ or ‘No’ as follows:
\begin{equation}
\label{eq:relevancy}
{p}_{j} = 
\begin{cases}
    p_{j}(Yes) & \text{if output Yes} \\
    \lambda~(1-p_{j}(No)) & \text{if output No}
\end{cases}
\end{equation}
where $\lambda$ is a small value that ensures ranking irrelevant~(answer “No”) below relevant evidences~(answer “Yes”). The probability $p_{j}$ is calculated with three different settings.

\textbf{\texttt{GAIS-ALL:}}~
We select Softmax normalization across all tokens in the dictionary and extract the probability of the generated token ($p(Yes)$ if the output is "Yes" and $1-p(No)$ if the output is "No"). The $P_{j}$ then will be calculated according to Eq.~\ref{eq:relevancy}.

\textbf{\texttt{GAIS-YN}:} This setting employs prompt-based classification which involves categorizing tokens into "Yes" and "No" classes, each with all existing variants of yes~(e.g Yes, yes, YES) and no~(e.g no, No, NO) token IDs in the dictionary. The Softmax is applied exclusively on the cumulative sum of the tokens within these two classes. The class with the highest probability determines the generated token, and its probability serves as the extracted score~$p_{j}$.

\textbf{\texttt{GAIS-YNO}:} As in the second approach, we categorize "Yes" and "No" tokens but also sum the probabilities of the remaining tokens for the class "other" for Softmax normalization. 
The final probability~${p}_{j}$ is calculated depending on the class with the highest probability, only considering the outputs for "Yes" and "No". 

Using one of the strategies above, we re-rank the candidate evidence based on~$p_j$ and select the top-K as the retrieved evidence samples~$\mathbb{C}''$ for fact verification.

\subsection{Fact Verification}
\label{sec:fact_verification}
Based on the retrieved text and image evidence samples, we suggest a fact verification component to predict the veracity of an input claim~(Figure~\ref{fig:pipeline}, right). 
We consider two models, \textit{Mistral}~\cite{DBLP:Mistral} for text-only evidence, and \textit{LLaVA}~\cite{DBLP:llava} for multimodal evidence.

For \textit{text-based misinformation detection}, we pair each retrieved evidence sentence with the input claim~$(Q, S_{j}),\ j \in [1,K]$.
For \textit{multimodal misinformation detection}, we first form a set of multimodal pairs~$\mathbb{P}$ by augmenting each retrieved sentence~$S_j$ with its corresponding images~$\mathbb{I}_{j}$ from the same document, and likewise, each retrieved image $I_j$ with its 
corresponding sentences in the text. Then, we create prompts using the claim and each of these multimodal pairs.

\subsubsection{Prompting}
\label{sec:fact_verification-prompting}
We follow two strategies to create a prompt template for fact verification: one-level prompting~(an example is shown in Figure~\ref{fig:pipeline}, right) and two-level prompting. 
In one-level prompting, models are instructed to answer 'Yes' if the evidence support the claim, 'No' if it refutes it and “None” if it does not provide enough information. This enables us to directly obtain all three labels from the model.
%
However, this strategy might lead to a bias towards 'Yes' or 'No' decision as the definition of NEI is not very explicit. 
Thus, in two-level prompting, the idea is to first explicitly ask the model if the evidence is enough to support or refute the claim to determine samples of “NEI” class. Then, we input all remaining samples with enough information~(answer “Yes”) to the second level and ask the model whether a sample supports or refutes the claim. In this way, the models are instructed to answer 'Yes' or ‘No’ in both levels.

\subsubsection{Classification}
We use prompt-based classification to ensure the model output is robust and we avoid hallucination. 
To this end, the normalization and extraction of the generated token are based on \texttt{GAIS-YN}, as mentioned in Section~\ref{ranking} for both prompting strategies. 
Note that for one-level prompting, as we have three classes~(Yes/No/None tokens), Softmax is applied across all of them.
Once we have the generated label for each claim-evidence pair~$(Q, P \in \mathbb{P}$), we apply majority voting 
to determine whether the claim is supported, refuted, or NEI. In case of equal number of maximum votes for multiple labels, we use the one with maximum extracted probability from the model. 











%
%
%
%
%
%

\section{Experiments}
\label{sec:experiments}

This section provides details on the experimental setup (Section~\ref{sec:exp_setup}), results for both evidence retrieval and fact verification tasks (Section~\ref{sec:exp_results} and~\ref{sec: fakeNewsDetection} respectively), and a generalization study on \textit{Factify} dataset (Section~\ref{sec:generalization}).

\subsection{Experimental Setup}
\label{sec:exp_setup}
In this section, we describe the setting of experiments including the multimodal datasets used in the experiments (Section~\ref{sec:datasets}), baseline methods (Section~\ref{sec:baselines}) and implementation details (Section~\ref{sec:implementation_details}). 
\subsubsection{Dataset} 
\label{sec:datasets}
To evaluate the performance and generalization across various domains of the proposed pipeline, we conduct experiments on two datasets, i.e., \textit{MOCHEG}~\cite{DBLP:mocheg} and \textit{Factify}~\cite{DBLP:factify}. 

\textit{MOCHEG} is a multimodal fact checking benchmark. 
This dataset is originally based on textual claims from \textit{Snopes} and \textit{PolitiFact}, with associated text and image evidences. It is divided into a train, 
validation and test set. Since we propose a zero-shot approach, we only use the 
validation set for hyperparameter selection and test set for evaluation. 
%

\textit{Factify} is a fact-checking benchmark featuring news from India and United States, with both claim and evidence being multimodal. The data is originally labeled in five categories of \textit{Support\_Text}, \textit{Support\_Multimodal}, \textit{Insufficient\_Text}, \textit{Insufficient\_Multimodal} and \textit{Refute}. In order to effectively apply our approach and compare it to the baselines, we convert it to the broad 3-class categories including \textit{Support}~(Support\_Multimodal \& Support\_Text), \textit{Insufficient}~(Insufficient\_Multimodal \& Insufficient\_Text) and \textit{Refute}.  Since the test set labels are not publicly available due to the dataset's origin from an open challenge, we utilize the development set for hyperparameter selection and the validation set for evaluation as proposed by~\cite{DBLP:logically}.
\subsubsection{Baselines}
\label{sec:baselines}
To compare our approach to the state of the art, we use two baselines. 
\textit{MOCHEG}~\cite{DBLP:mocheg} is the first 
end-to-end multimodal fact-checking method which considers both tasks of evidence retrieval and verification. 
As mentioned in the Section~\ref{sec:related_work}, it fine-tunes SBERT model~\cite{DBLP:SBERT} for text retrieval and CLIP~\cite{DBLP:CLIP} for image retrieval in a contrastive manner and based on cosine similarity between the input claim and candidate evidence. 
For the verification task, it uses the CLIP model to encode both claim and evidence and then an attention layer to compute the distribution between them. 
The combined representation is subsequently passed through a classification layer to predict the final labels. 
We have used the official implementation of \textit{MOCHEG} provided on GitHub\footnote{https://github.com/VT-NLP/Mocheg} with the best parameters reported.

We also compare the approaches to \textit{Logically}~\cite{DBLP:logically} on \textit{Factify} validation set as this method was ranked first on the challenge leaderboard. \textit{Logically} treats the challenge as a multimodal entailment task and proposes an ensemble model which combines predictions of two uni-modal models (a transformer architecture for text and a ResNet-50~\cite{DBLP:ResNet} for image) fine-tuned on the task dataset. As \textit{Logically} is fully-supervised and trained specifically on this dataset, it has an advantage over our proposed approach, which operates in a zero-shot setting.

\subsubsection{Implementation Details}
\label{sec:implementation_details}
For the initial retriever (Section~\ref{sec:initial retriever}), the input length of claims and textual evidences 
is truncated to 77 tokens and evidence images are resized to  $224\times224\times3$ pixels.
We use Mistral-OpenOrca~(7B)\footnote{https://huggingface.co/Open-Orca/Mistral-7B-OpenOrca}~\cite{DBLP:Mistral} for text retrieval and fact verification with text evidence due to its demonstrated superior performance over many advanced LLMs, such as LLaMA2, and its exceptional capabilities across various natural language processing tasks~\cite{DBLP:Mistral}. For image retrieval, we use InstructBLIP~(Flan-T5-xl)\footnote{https://huggingface.co/docs/transformers/en/model\_doc/instructblip}~\cite{DBLP:Instruct-blip} as it has ranked first in SEED-Bench~\cite{DBLP:seed-bench1} leaderboard by achieving the best performance based on the averaged results across nine evaluation dimensions~\cite{DBLP:seed-bench1}. 
Finally, we use the LLaVA v1.6-Mistral~(7B)\footnote{https://huggingface.co/llava-hf/llava-v1.6-mistral-7b-hf}~\cite{llava-mistral} model for fact verification with multimodal evidence. LLaVA was among the top-ranked models on the SEED-Bench leaderboard for question answering tasks~\cite{DBLP:seed-bench2}. Additionally, since LLaVA v1.6-Mistral utilizes the same language model (Mistral), it makes fact verification with text and multimodal evidence comparable. 
All models are loaded in 8-bit quantization with a maximum token length of 2048 for Mistral and LLaVA and 512 for InstructBLIP.
All experiments were conducted on two NVIDIA A3090, 24-GB GPUs. We set the size of the initial retriever $N=100$ and choose the best working prompt according to Section~\ref{sec: results_prompt} for the experiments. We will make our code publicly available for further research.

\subsection{Results for Evidence Retrieval}
\label{sec:exp_results}
In this section, we report and discuss the performance of proposed pipeline in comparison to state-of-the-art baseline on evidence retrieval using \textit{MOCHEG} dataset~(Section~\ref{sec: evidence retrieval}) and annotated data (Section~\ref{sec: results_annotated}) that  addresses 
the issue of incomplete ground truth evidences in the retrieval step. 
We then discuss the impact of different prompts (Section~\ref{sec: results_prompt}) and re-ranking strategies~(Section~\ref{sec: results_settings}) for evidence retrieval.

\subsubsection{Results on MOCHEG}
\label{sec: evidence retrieval}
Similar to \textit{MOCHEG}, for each claim, we evaluate the retrieval performance based on precision, recall and mean average precision~(mAP) scores using the top-$K$ retrieved text and image evidences. 
The results on the test set of \textit{MOCHEG} with $K \in {\{1,2,5,10}\}$ are shown in Table~\ref{table:retrieval}.
\begin{table}[t]
\small
\centering
\caption{Precision~(Pre), Recall~(Rec) and mean Average Precision (mAP) of models for evidence retrieval on MOCHEG dataset with N=100 in percent [\%].}
\label{table:retrieval}
\begin{tabularx}{\linewidth}{lcYYYYYY}
\toprule
\multirow{2}{*}{\bf{Method}} & \multirow{2}{*}{\bf{K}} & \multicolumn{3}{c}{\bf{Text Retrieval}} & \multicolumn{3}{c}{\bf{Image Retrieval}}\\ \cmidrule(lr){3-5} \cmidrule(lr){6-8}
& & \bf{$Pre$} & \bf{$Rec$} & \bf{$mAP$} & \bf{$Pre$} & \bf{$Rec$} &\bf{$mAP$}\\
\midrule
\multirow{4}{*}{\bf{MOCHEG}} & 1 & 27.14 & 8.63 &  \bf{27.14}  &  8.56  & 6.46  &   8.56 \\ &  2 & 21.81 & 13.13 & \bf{20.39} & 6.89  &  10.01  &  8.91 \\ &  5   & 13.70  &  18.77   &  \bf{16.23}   &  4.18   & 15.40  & 10.17 \\   & 10  &   9.18    & 23.59  &  \bf{15.79}  &  2.73  &  19.95   &  10.89\\
\midrule
\multirow{4}{*}{\bf{\modelName}} & 1 & 26.34 & 7.44 &  26.34  &  9.65  & 7.13  &   \bf{9.65} \\ & 2 & 20.64 & 10.68 & 19.00 & 7.29  &  10.18  &  \bf{9.50} \\ & 5   & 14.10  &  16.40  &  14.77   &  4.19   & 14.80  & \bf{10.42} \\ &  10  &   9.58    & 21.04  &  13.97  &  2.54  &  18.01   &  \bf{10.96} \\   
\bottomrule
\end{tabularx}
\end{table}

The results indicate that mAP decreases by increasing $K$ which is expected since a larger number of retrieved evidences may lead to the inclusion of noisy or redundant documents, reducing the precision at higher ranks. 
Besides, for example in image retrieval, the average number of gold evidences in the retrieved top-100 is 1.26 which means most of the claims have only one labeled gold image evidence.
The results also demonstrate that \modelName~outperforms the \textit{MOCHEG} even in zero-shot setting for image retrieval highliting its ability to retrieve images with higher contextual relevance. For text retrieval,~\modelName~is comparable but falls slightly below the baseline. However, we believe that the assessment of evidence retrieval may not accurately reflect the true capabilities of the models under scrutiny as not all relevant pieces of evidences are appropriately labeled. This is because the ground truth only considers evidence used in \textit{Snopes} and \textit{PolitiFact}. But the large evidence collection might contain more relevant evidences that are disregarded.
In the following section, we investigate this issue in more detail. 

\subsubsection{Results on Annotated Data}
\label{sec: results_annotated}
\begin{figure*}[t]
    \centering
  \includegraphics[width=\linewidth]{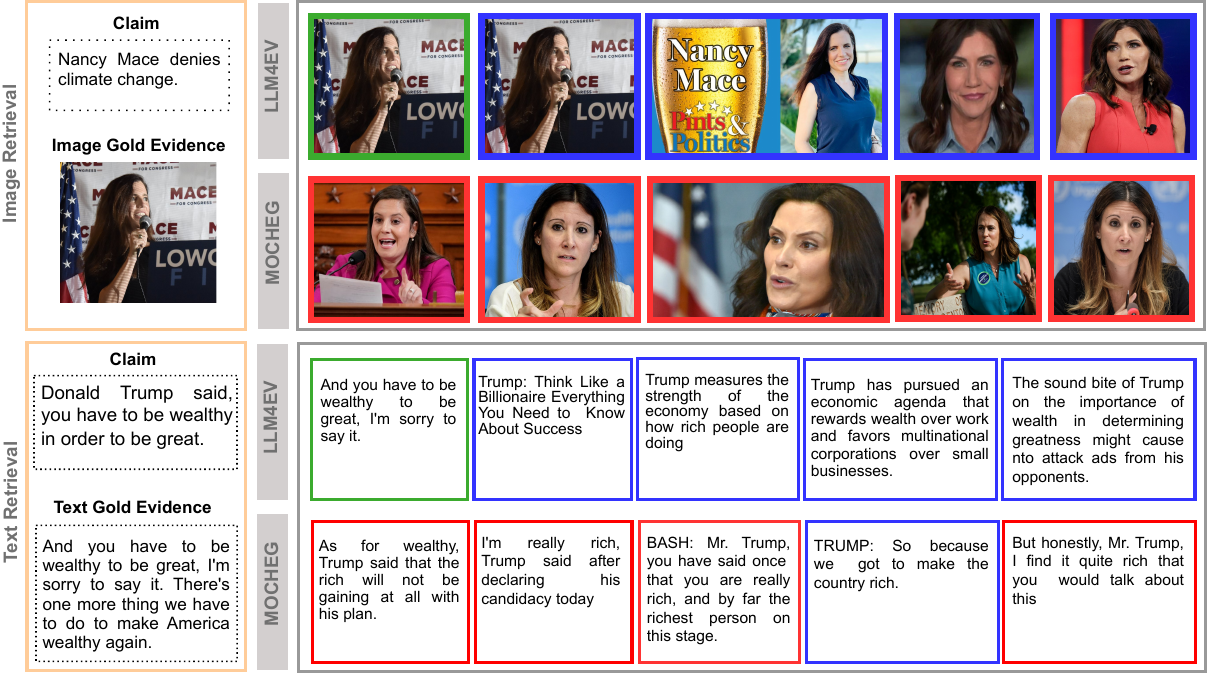}
  \caption{Qualitative example for evidence retrieval component which also shows the incomplete ground truth issue in the \textit{MOCHEG} dataset. Green border shows the labeled ground truth, Blue border shows evidences with similar content to the ground truth which are left unlabeled and red border shows irrelevant content.}
  \label{fig3}
\end{figure*}

While \textit{MOCHEG}~\cite{DBLP:mocheg} is one of the pioneering datasets for evaluating multimodal evidence retrieval, 
a significant challenge arises from incomplete annotation of relevant ground truth evidence samples. 
For example, in image retrieval each claim is associated with a set of labeled images meant to serve as ground-truth image evidence, while there exist other visually similar images within the dataset that are left unlabeled~(Figure~\ref{fig3}).
So the results can be misleading as the actual relevant evidences sometimes are not considered due to missing labels. 
Also, some relevant images are indeed relevant but sometimes do not serve as direct evidence to the claim. This is also the case for text retrieval as shown in Figure~\ref{fig3}. This inconsistency undermines the integrity of the dataset, introduces ambiguity and noise 
and leads to a diminished performance in identifying relevant evidence samples.  

To address these aforementioned issues in multimodal evidence retrieval and ensure fair and accurate assessments of system performance, we provide 
a subset of \textit{MOCHEG} test set with more complete and accurate annotations for both text and image evidence retrieval. 
The idea is to ensure all the top-10 relevant candidate evidences are labeled for the sampled claims by considering both baselines. 
For each modality, we consider three evidence sets: (1)~the top-10 evidences~$\mathbb{C}'$ from the initial retrieval using \textit{MOCHEG}~(\textit{Initial-top10}), (2)~the re-ranked evidences~$\mathbb{C}''$~(\textit{Re-ranked-top10}) and (3)~their union~$\mathbb{C}' \cup \mathbb{C}''$~(\textit{Union}). Please note that we only retain distinct candidate evidence when there is overlap for a claim.

\begin{figure*}[ht]
    \centering
  \includegraphics[width=\linewidth]{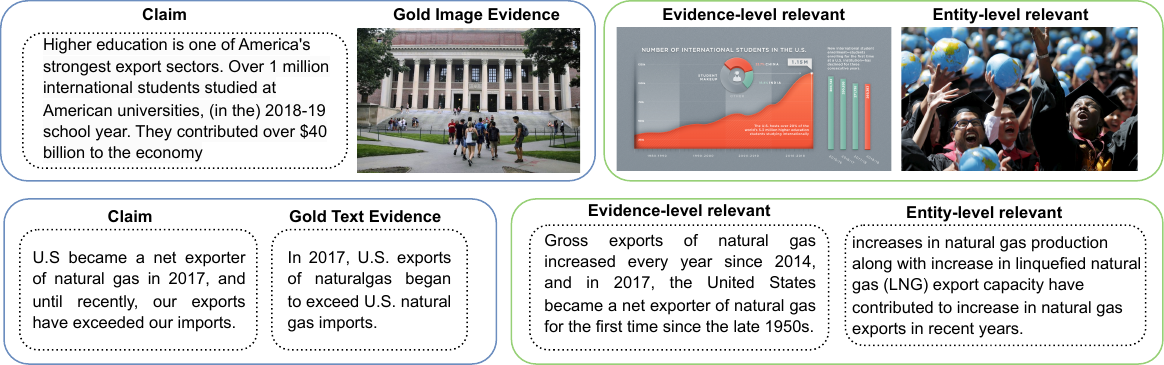}
  \caption{An example of data annotation at entity level and evidence level for image retrieval (top) and text retrieval (bottom).}
  \label{fig4}
\end{figure*}

\paragraph{Annotation Process}
For a fair assessment of the evidence retrieval, an expert manually performed annotations for the \textit{Initial-top10} and \textit{Re-ranked-top10} using ten randomly selected queries from the \textit{MOCHEG} test dataset. Note that the ground-truth labels for the \textit{Union} set can be derived from these annotations.
The annotation process is performed on three levels. (1)~Entity-level: This level denotes when the candidate evidence is relevant but solely shares the same entity or topic without offering direct support or refutation of the claims.
(2)~Evidence-level: This is the case if the candidate provides direct evidence to support or reject the claim.
(3)~Overall: If either of the preceding two levels is deemed true, we designate the total relevancy as true.
Figure~\ref{fig4} demonstrates an example of the annotation.



\paragraph{Results}
We evaluate the performance of models on the annotated union set while considering the "Overall" labels for each claim. Results for other two types of annotated labels~(evidence-level and entity-level) will be additionally provided on GitHub but they generally lead to the same conclusions. The result are shown in Table~\ref{table:annotated}.
\begin{table}[t]
\small
\centering
\caption{Precision~(Pre), Recall~(Rec) and mean Average Precision (mAP) of models for evidence retrieval on union annotated set in percent~[\%]. Mistral-Open-Orca and InstructBlip models are used for text and image retrieval respectively.}
\label{table:annotated}
\begin{tabularx}\linewidth{lcYYYYYY}
\toprule
\multirow{2}{*}{\bf{Method}} & \multirow{2}{*}{\bf{K}} & \multicolumn{3}{c}{\bf{Text Retrieval}} & \multicolumn{3}{c}{\bf{Image Retrieval}}\\ \cmidrule(lr){3-5} \cmidrule(lr){6-8} & & \bf{$Pre$} & \bf{$Rec$} & \bf{$mAP$} & \bf{$Pre$} & \bf{$Rec$} &\bf{$mAP$}\\
\midrule
\multirow{3}{*}{\bf{MOCHEG}} & 1 & 80.00 & 19.96 &  80.00  &  80.00  & 34.02  &   80.00 \\  & 5 & 64.00 & 45.17 & 64.23 & 56.00  &  67.05  &  72.07 \\ & 10   & 47.00  &  57.83  &  58.40  & 43.00   & 89.27  & 75.29\\
\midrule
\multirow{3}{*}{\bf{\modelName}} & 1 & 90.00 & 20.79 &  \bf{90.00}  &  90.00  & 35.45  &   \bf{90.00} \\  & 5 & 70.00 & 42.46 & \bf{69.80} & 56.00  &  67.05  &  \bf{72.77} \\ & 10   & 60.00  &  76.75  &  \bf{70.48}   &  43.00   & 89.27  & \bf{75.83}\\

\bottomrule
\end{tabularx}
\end{table}
It clearly shows that \modelName~has a significant improvement over the baseline in both image and text retrieval using the revised labels. This confirms our hypothesis that the model retrieves more relevant pieces of evidence compared to the baseline, as illustrated in Figure~\ref{fig3}. However, this improvement is not reflected in the numbers for the \textit{MOCHEG} test set.

\subsubsection{Prompt Impact}
\label{sec: results_prompt}
As mentioned in Section~\ref{sec:prompting}, we also experiment with different prompts to investigate the performance for both image and text retrieval. 
We report results for the prompts shown in Table~\ref{tab:prompts_resutls} using the following template:
\begin{itemize}
    \item Text prompt for \textit{Mistral}: \newline 
    \texttt{[prompt] \textbackslash n \newline
    \#\#\#~Query: [${Q}_{i}$]\textbackslash n \newline
    \#\#\#~ corpus: [${S}_{j}$]\textbackslash n\newline
    \#\#\#~Answer:}
    \item Image prompt for \textit{InstructBLIP}: \newline 
    \texttt{[prompt]\textbackslash n \newline
    \#\#\#~Query: [${Q}_{i}$]}
\end{itemize}
\begin{table}[t] 
\centering
\small
\caption{Results of \modelName~for both text and image retrieval with different prompts. Best results are highlighted in bold.}
\label{tab:prompts_resutls}
\begin{tabularx}{\linewidth}{llcYYY}
\toprule
\bf{Media} & \bf{Prompts} & \bf{K} & \bf{Pre} & \bf{Rec} & \bf{mAP}\\
\midrule
\multirow{3}{*}{\bf{txt}} & \multirow{3}{*}{\makecell[l]{Is this corpus related to the\\ query? Answer with yes\\ or no.}} & 1 & 26.19 & 7.59 &  \bf{26.19} \\ & &  5 & 14.62 &  18.31 & \bf{15.82} \\  & & 10 & 10.81 & 24.60 & \bf{15.40} \\
\midrule
\multirow{3}{*}{\bf{txt}} & \multirow{3}{*}{\makecell[l]{Is query and corpus mentioning\\ the same person or topic?\\ Answer with yes or no.}} & 1 & 19.94 & 5.32 &  19.94 \\ & &  5 & 10.81 &  13.09 & 11.23 \\  & & 10 & 7.71 & 17.91 & 10.78 \\
\midrule
\multirow{3}{*}{\bf{txt}} & \multirow{3}{*}{\makecell[l]{Is this corpus an evidence\\ for the query? Answer with\\ yes or no.}} & 1 & 21.39 & 5.83 &  21.39 \\ & &  5 & 17.84 &  9.22 & 16.08 \\  & & 10 & 7.91 & 17.95 & 11.58 \\
\hhline{======}
\multirow{3}{*}{\bf{img}} & \multirow{3}{*}{\makecell[l]{Does this query describe\\ the image?}} & 1 & 7.71 & 5.36 &  7.18 \\ & &  5 & 3.66 &  13.27 & 8.54 \\  & & 10 & 2.54 & 17.92 & 9.22 \\
\midrule
\multirow{3}{*}{\bf{img}} & \multirow{3}{*}{\makecell[l]{Based on the query below,\\ is it related to the image?}} & 1 & 6.89 & 5.20 &  6.89 \\ & &  5 & 3.73 &  13.47 & 8.46 \\  & & 10 & 2.59 & 18.90 & 9.26 \\
\midrule
\multirow{3}{*}{\bf{img}} & \multirow{3}{*}{\makecell[l]{Is this image and text query\\ mentioning the same person\\ or topic?}} & 1 & 9.65 & 7.13 &  \bf{9.65} \\ & &  5 & 4.19 &  14.80 & \bf{10.42} \\  & & 10 & 2.54 & 18.16 & \bf{10.96} \\
\bottomrule
\end{tabularx}
\end{table}

For image retrieval, the prompt asking whether the image and text mentions the same person
or topic performs the best. This is because most gold image evidence in the dataset shares the same entity or topic rather than providing direct evidence for the claim (only 1.5\% of images were identified as evidence-level in the annotated data). While other similar prompts may also be useful in practice, we use this one for our experiments due to its superior results.
% ...
In text retrieval, prompt that directly assess the relevance of the corpus to the claim demonstrate better performance compared to others 
as text evidence usually carries key relevancy information and serve as the evidence to the claim. However, given that documents are segmented into sentences and evidence typically comes from the comprehension of long paragraphs rather than single sentence, the prompt directly asking if a corpus is an evidence for the claim performs worse than the previous one.
\subsubsection{Impact of Re-ranking Strategies}
\label{sec: results_settings}
\begin{figure}
\centering
\begin{subfigure}[b]{\linewidth}
\centering
\includegraphics[width=\textwidth]{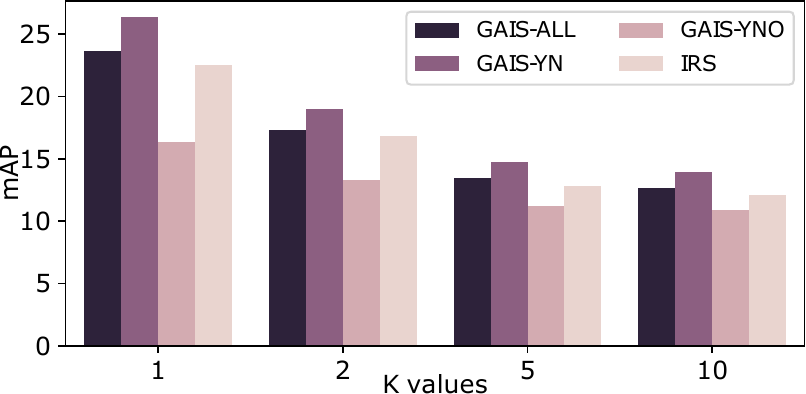}
\label{fig:settings_txt}
\end{subfigure}

\begin{subfigure}[b]{\linewidth}
\centering
\includegraphics[width=\textwidth]{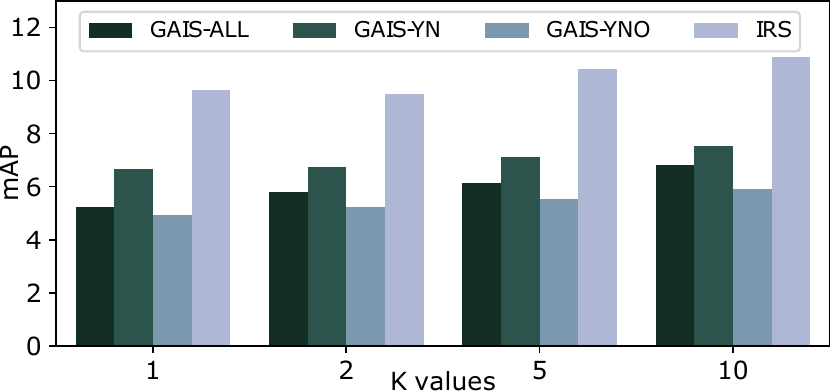}
\label{fig:settings_img}
\end{subfigure}
\caption{Mean average precision (mAP) of \modelName~using different re-ranking strategies~(Section~\ref{ranking}) for top: text retrieval and bottom: image retrieval.}
\label{fig:both_subfigures}
\end{figure}
Figure~\ref{fig:both_subfigures} illustrates a comparison of the 
mAP for various re-ranking strategies, as referenced in Section~\ref{ranking}. The results indicate that the CLIP model score proves superior for image retrieval, while the \texttt{GAIS-YN} scores from LLM outperforms others in text retrieval. This discrepancy is likely attributed to the inherent hallucination issue associated with LLMs, wherein the model may generate outputs that do not necessarily match the intended prompts. Accordingly, the model may assign probabilities to irrelevant tokens that are not part of the correct answer sets. Incorporating these redundant probabilities during the normalization process~(as in \texttt{GAIS-All} and \texttt{GAI-YNO}) introduces noise, thereby affecting the final score and degrading the final ranking performance.

\subsection{Results for Fact Verification}
\label{sec: fakeNewsDetection}
\begin{table*}[t]
\centering
\setlength{\tabcolsep}{4pt}
\caption{Precision~(Pre), Recall~(Rec) and F1-score~(F1) of models for fact verification task on MOCHEG test set. Mistral-Open-Orca model is used for text evidence and LLaVA model is used for multimodal evidence. All retrieved evidences are with K=5}
\label{table:fndetection_mocheg}
\begin{tabularx}{\linewidth}{lccYYYYYYYYYY}
\toprule
\multirow{2}{*}{\bf{Method}} & \multirow{2}{*}{\bf{Evidence}} & \multirow{2}{*}{\bf{Modality}} & \multicolumn{3}{c}{\bf{Supported}} & \multicolumn{3}{c}{\bf{Refuted}} &  \multicolumn{3}{c}{\bf{NEI}} & \multirow{2}{*}{\makecell[c]{\bf{micro}\\\bf{F1}}} \\
\cmidrule(lr){4-6}
\cmidrule(lr){7-9}
\cmidrule(lr){10-12} 
& & & \bf{$Pre$} & \bf{$Rec$} & \bf{$F1$} & \bf{$Pre$} & \bf{$Rec$} & \bf{$F1$} & \bf{$Pre$} & \bf{$Rec$} & \bf{$F1$} & \\
\midrule
\multirow{2}{*}{\bf{\FNmodelName}} & Retrieved & text & 0.639 & 0.380 &  0.477 & 0.393 & 0.896 &  0.547 & 0.352 &  0.044 & 0.079 & 0.440 \\ 
& Gold & text & 0.500 & 0.677 & 0.575 & 0.816 & 0.406 & 0.542 & 0.411 & 0.472 & 0.439 & 0.518\\
\midrule
\multirow{2}{*}{\bf{\FNmodelName}} & Retrieved & multimodal & 0.622 & 0.491 & 0.549 & 0.460 & 0.398 &  0.428 & 0.451 & 0.316 & 0.372 & 0.451\\ 
& Gold & multimodal & 0.453 & 0.800 & 0.578 & 0.858 & 0.426 & 0.569 &  0.422 & 0.501 & 0.457 & \bf{0.534} \\
\midrule
\bf{MOCHEG} & Gold & multimodal & 0.502 & 0.479 & 0.490 & 0.481 & 0.811 & 0.604 & 0.533 & 0.191 &  0.282  & 0.491 \\
\bottomrule
\end{tabularx}
\end{table*}

\begin{table*}[t]
\centering
\setlength{\tabcolsep}{4pt}
\caption{Precision~(Pre), Recall~(Rec) and F1-score~(F1) of models for fact checking task on Factify validation set. Mistral-Open-Orca model is used for text evidence and LLaVA model is used for multimodal evidence. All retrieved evidences are with K=5}
\label{table:fndetection_factify}
\begin{tabularx}{\linewidth}{lccYYYYYYYYYY}
\toprule
\multirow{2}{*}{\bf{Method}} & \multirow{2}{*}{\bf{Evidence}} & \multirow{2}{*}{\bf{Modality}} & \multicolumn{3}{c}{\bf{Supported}} & \multicolumn{3}{c}{\bf{Refuted}} &  \multicolumn{3}{c}{\bf{NEI}}  & 
\multirow{2}{*}{\makecell[c]{\bf{micro}\\\bf{F1}}}\\
\cmidrule(lr){4-6}
\cmidrule(lr){7-9}
\cmidrule(lr){10-12}
& & & \bf{$Pre$} & \bf{$Rec$} & \bf{$F1$} & \bf{$Pre$} & \bf{$Rec$} &\bf{$F1$} & \bf{$Pre$} & \bf{$Rec$} &\bf{$F1$} & \\
\midrule
\multirow{2}{*}{\bf{\FNmodelName}} & Retrieved & text & 0.533 & 0.634 &  0.579 &   0.507 &  0.620 & 0.558  & 0.492 & 0.344 & 0.405 & 0.515\\ & Gold & text & 0.578 & 0.608 & 0.593 & 0.455 & 0.806 & 0.581 & 0.778 & 0.438 & 0.560 & 0.580\\
\midrule
\multirow{2}{*}{\bf{\FNmodelName}} & Retrieved & multimodal & 0.636 & 0.446 & 0.524 & 0.489 & 0.696 &  0.575 & 0.495 & 0.480 & 0.487 & 0.521\\ & Gold & multimodal & 0.636 & 0.726 & 0.678 & 0.485 & 0.804 & 0.605 & 0.521 & 0.496 & 0.508 & \underline{0.595} \\
\midrule
\bf{MOCHEG} & Gold & multimodal & 0.454 & 0.688 & 0.547 & 0.578 & 0.670 & 0.621 & 0.475 &  0.193 & 0.275    &   0.456 \\
\midrule
\bf{Logically} & Gold & multimodal & 0.830 & 0.860 & 0.850 & 1.00 & 1.00 & 1.00 & 0.850 &  0.830 & 0.840   &   \textbf{0.870} \\
\bottomrule
\end{tabularx}
\end{table*}
Table~\ref{table:fndetection_mocheg} presents the performance of various models across the \textit{MOCHEG} test set on fact verification task based on micro F1-score (F1). 
To evaluate the impact of each type of evidence for classification, we design ablated models by considering the text evidence only or multimodal evidence. We have used one-level prompting for text evidence and two-level prompting for multimodal evidence as mentioned in Section~\ref{sec:fact_verification-prompting}
Our findings reveal that fact verification utilizing multimodal gold evidences outperformed its text-only counterparts. 
This suggests that while image evidences alone may only share the same entity or topic and have less impact in the final verification task, together with text, they contain a wealth of information and serve as more concise indicators within the dataset. For example, claims that start with phrases like \textit{“A photograph of…”} or \textit{“The image shows…”} are multimodal, meaning an important part of the information is contained within the image modality. Therefore, fact verification based on multimodal pairs provides more significant gain than text only.
The results also show the various model performances in comparison to the state of the art. Our proposed approach achieves an improvement over the \textit{MOCHEG} baseline in the overall F1-score, surpassing it by 
4.3\% with respect to the multimodal gold evidences, despite operating in zero-shot manner. This suggests the potential of proposed method for verification task compared to \textit{MOCHEG} method which relies on fine-tuning for this specific dataset and incorporates a dense information retrieval component. One explanation for this superior performance is that the instruction-tuning data of LLaVA contains totally 158K samples and covers a wide range of multimodal tasks, including question answering and visual reasoning data, which allows it to effectively capture semantic relation between images and text.
After conducting an error analysis of the fact verification outcomes, 
we have observed that the majority of failure cases are due to a model's maximum length constraint. 
As we need to input the prompt template, claim, and complete textual evidence into the model, this might result in loss of content, particularly with lengthy evidence passages. 
We believe that an increase of the model's maximum length would boost the results, although this adjustment would also cause longer run times. Another type of failure occurs with multimodal claims where the evidence is in a chart image. This demonstrates that LLaVA struggles with text recognition and information extraction from charts, as mentioned in~\cite{DBLP:seed-bench1} as a common issue within similar models.
\subsection{Generalization Study}
\label{sec:generalization}
We conduct evaluations on the \textit{Factify} dataset to analyze the robustness and generalizability of employed methods in a cross-domain context 
which is crucial for real-world misinformation detection. Misinformation varies widely across platforms, topics, and languages and a model that excels only on specific data may fail with new examples. 

Table~\ref{table:fndetection_factify} presents the performance of various models across the \textit{Factify} validation set on fact verification task with different types of evidence.
The results demonstrate that while the performance of proposed method may fall short of \textit{Logically}, which is fine-tuned specifically for the dataset, it notably surpasses \textit{MOCHEG} that has been trained on a different benchmark and fails to generalize to the new domains, topics, and characteristics in \textit{Factify}. 
%
This indicates the robustness and efficacy of our approach in adapting to unseen data domains without the need for extensive fine-tuning.%

\section{Conclusions}
\label{sec:conclusion}
%
In this paper, we have presented a pipeline for multimodal misinformation detection, introducing a novel re-ranking approach for evidence retrieval using both LLMs and LVLMs~(\modelName). The claim and retrieved evidence samples (texts and images) serve as the input for LVLM-based fact verification~(\FNmodelName). 
We demonstrated the capabilities of LLMs and LVLMs for evidence retrieval and fact verification in zero-shot settings, employing them as rankers for text and image retrieval and used a prompting strategy which allows us to extract ranking scores and determine the veracity of claims. 
%
Furthermore, we tackled the issue of incomplete and inaccurate ground-truth labels for evidence retrieval tasks and provided an improved annotation to enable a more meaningful system evaluation. 
%
Our experiments on two datasets have demonstrated that our proposed zero-shot approach outperforms a supervised baseline for multimodal misinformation detection. Moreover, it has shown much better generalization capabilities as it is not fine-tuned on relatively small domain- and topic-specific datasets. 
%

In future work, we will focus on the interpretability of the pipeline. 
We aim to achieve this by obtaining explanations directly from the model itself, enabling us to discern the specific contributions of different evidence components towards the final decision-making process. 
Furthermore, 
Given the lack of high-quality annotations in current multimodal datasets, a comprehensive, well-annotated, large-scale dataset is crucial for advancing research in multimodal evidence retrieval for misinformation detection. Finally, fine-tuning suitable L(V)LMs for evidence retrieval and fact verification can further improve performance. 
\begin{acks}
This work was partially funded by the German Federal Ministry of Education
and Research (BMBF, FakeNarratives project, no. 16KIS1517).
\end{acks}
\bibliographystyle{ACM-Reference-Format}
\bibliography{main}

\end{document}